\documentclass[10pt,letterpaper,twocolumn]{article}
\usepackage[letterpaper,left=.75in,right=.75in,top=.75in,textheight=9in]{geometry}
\usepackage[no-math]{fontspec}
\usepackage{amsmath,amssymb,graphicx,booktabs,multirow}
\usepackage{titlesec,enumitem}
\usepackage[font=normalsize,labelsep=colon,skip=5pt]{caption}
\usepackage[numbers,sort&compress]{natbib}
\usepackage[hyphens]{url}
\usepackage[hidelinks]{hyperref}
\usepackage[capitalize]{cleveref}
\crefname{section}{Sec.}{Secs.}
\Crefname{section}{Section}{Sections}
\crefname{table}{Tab.}{Tabs.}
\Crefname{table}{Table}{Tables}
\newcommand{\Sc}{S^{\mathrm{c}}}
\hypersetup{pdftitle={Seeing vs. Knowing: Counterfactual Supervision and Correction for 3D-Aware Object Editing},pdfauthor={Jian Zhang and Zhijun Zhang}}

\makeatletter
\renewcommand{\normalsize}{\@setfontsize\normalsize{10pt}{11pt}%
  \abovedisplayskip=7pt plus 2pt minus 3pt
  \belowdisplayskip=\abovedisplayskip
  \abovedisplayshortskip=2pt plus 2pt
  \belowdisplayshortskip=4pt plus 2pt minus 1pt}
\renewcommand{\small}{\@setfontsize\small{9pt}{10pt}}
\makeatother
\normalsize
\titleformat{\section}{\centering\fontsize{12}{14}\selectfont\bfseries}{\thesection}{.5em}{}
\titleformat{\subsection}{\raggedright\fontsize{11}{12}\selectfont\bfseries}{\thesubsection}{.5em}{}
\titlespacing*{\section}{0pt}{2ex plus .5ex minus .2ex}{3pt plus 2pt minus 1pt}
\titlespacing*{\subsection}{0pt}{2ex plus .5ex minus .2ex}{3pt plus 2pt minus 1pt}
\setlist[itemize]{after=\vspace{-3pt},leftmargin=13pt,topsep=3pt plus 1pt,itemsep=1pt plus .5pt,parsep=0pt}
\AddToHook{selectfont}{\baselineskip=\dimexpr\baselineskip\relax plus .35pt minus .1pt\relax}
\renewenvironment{abstract}{\par\begingroup
  \centerline{\bfseries Abstract}\vspace{.5ex}
  \leftskip=10pt\rightskip=10pt\small\noindent\ignorespaces
}{\par\endgroup\vspace{1ex}}

\begin{document}
\twocolumn[{\centering
  \vspace*{.48in}
  {\fontsize{14.4}{16}\selectfont\bfseries
    Seeing vs.\ Knowing: Counterfactual Supervision and Correction\\
    for 3D-Aware Object Editing\par}
  \vspace{12pt}
  {\fontsize{12}{14}\selectfont\bfseries
    Jian Zhang\textsuperscript{1}, Zhijun Zhang\textsuperscript{1}\par}
  \vspace{3pt}
  {\normalsize\textsuperscript{1}School of Electronic and Information Engineering, South China University of Technology, Guangzhou, China\par
    202411084134@mail.scut.edu.cn, auzjzhang@scut.edu.cn\par}
  \vspace{12pt}
  \begin{minipage}{\textwidth}
    \captionsetup{type=figure}\centering
    \includegraphics[width=\textwidth]{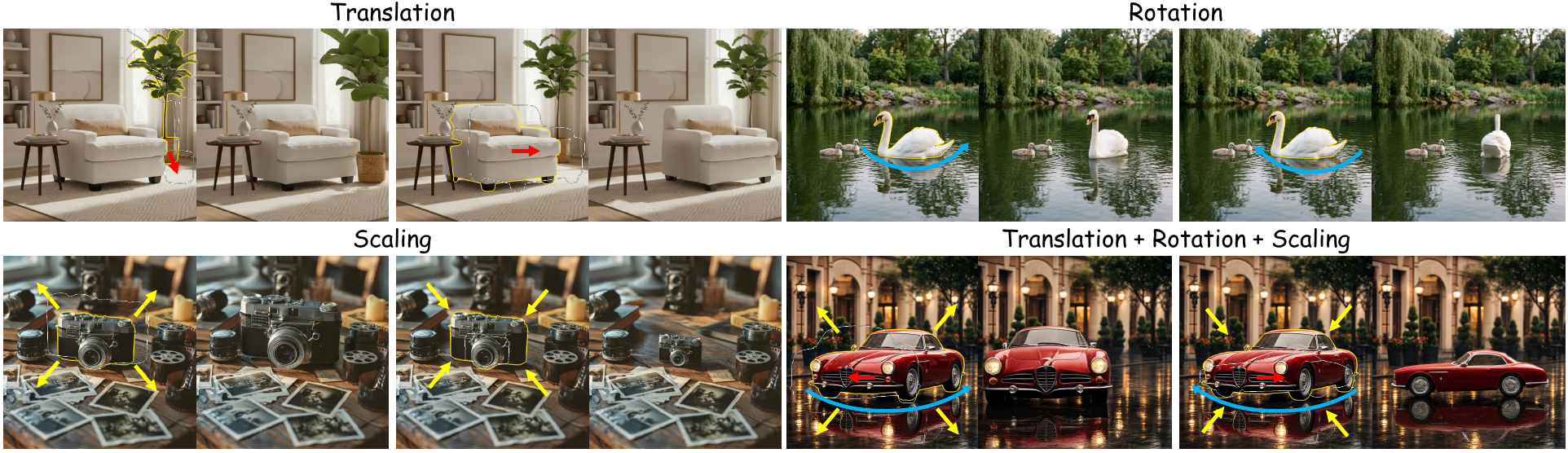}
    \caption{Our method performs translation, rotation, scaling, and their combinations across diverse scenes, while preserving background content and maintaining consistent object appearance, occlusions, shadows, and reflections.}
    \label{fig:demo}
\end{minipage}
\par\vspace{4pt}
}]
\begin{abstract}
Rotating, translating, or rescaling an object in a photograph requires revealing unseen surfaces, preserving visible appearance, and updating shadows and reflections.  Reconstructed 3D assets, rendered at the target pose as a 3D state, provide explicit geometric guidance, but their imperfect textures may conflict with the photograph, while standard paired training does not explicitly teach the editor which source to trust. 
To address this, we propose \textbf{counterfactual state supervision}, which constructs controlled photo--state conflicts through hidden-surface repainting, hidden geometry with re-rendered shadows and reflections, and visible-surface texture replacement.
This teaches the editor to follow the 3D state for unseen content while preserving photograph-visible appearance. However, learning this distinction does not ensure that it is maintained during sampling. We therefore introduce \textbf{symmetric counterfactual correction}, which uses state-side and photo-side twins, which share the photograph and the state respectively, to construct directed latent perturbations toward condition-inconsistent targets and train recovery toward the target specified by the current condition. Both strategies operate only during training, requiring no architectural changes or additional inference steps. Experiments on OBJect and SpatialEdit-Bench demonstrate strong spatial editing performance and content preservation, while controlled ablations show improved source selection and sustained responsiveness to conditioning during sampling.
\end{abstract}

\section{Introduction}
\label{sec:intro}

\begin{figure*}[t]
  \centering
  \includegraphics[width=\textwidth,height=9.5cm]{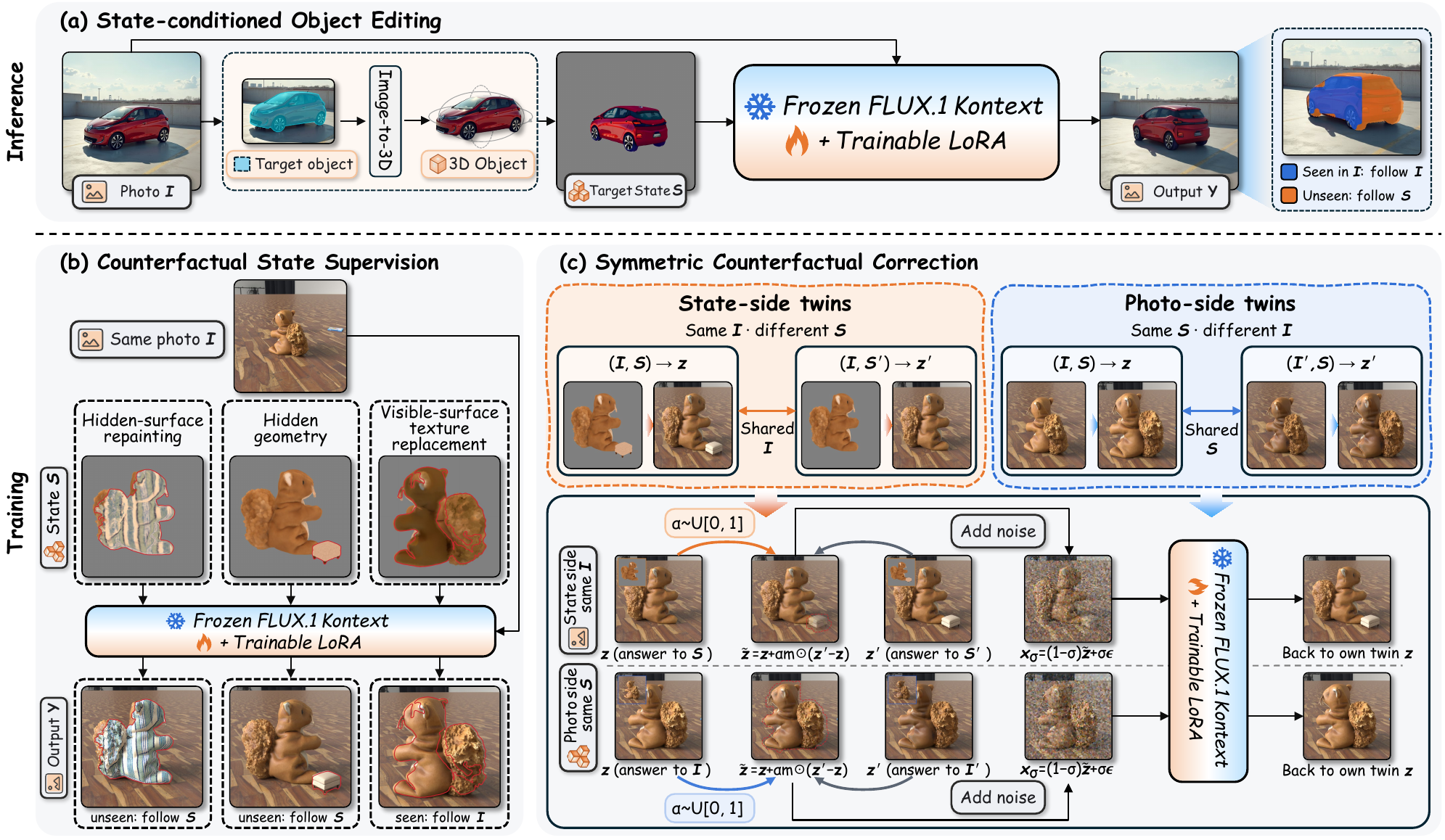}
    \caption{\textbf{Overview.}  
    (a)~A reconstructed 3D state guides editing: visible appearance follows the photograph, unseen content follows the state, and shadows and reflections adapt to the edited geometry.
    (b)~Counterfactual state supervision trains the editor with standard flow matching on counterfactual editing pairs constructed through hidden-surface repainting, hidden geometry, and visible-surface texture replacement.
    (c)~Symmetric counterfactual correction uses state-side twins (shared photograph) and photo-side twins (shared state) to train recovery from condition-inconsistent latent perturbations.
    }
  \label{fig:overview}
  \vspace{-10pt}
\end{figure*}

Rotating, translating, or rescaling an object in a photograph requires more than moving its visible pixels. 
Such edits can reveal previously unseen surfaces and require corresponding updates to occlusions, contact, shadows, and reflections, as illustrated in \cref{fig:demo}. Existing editors often rely on generative priors for unseen content and scene effects, which can lead to inconsistent appearances across edits~\cite{thinkinginboxes,object3dit,spatialedit,objectmover}. Recent image-to-3D models~\cite{hunyuan3d21,trellis,hunyuan3d2} offer an alternative by enabling textured 3D assets to be reconstructed from single photographs, registered to the scene, and rendered at target poses. We refer to their colour and normal renderings as a 3D state, which provides explicit geometric guidance and a consistent appearance reference for object editing (\cref{fig:overview}a).

However, a reconstructed state is not equally reliable everywhere. Its texture may differ from the appearance of photograph-visible surfaces~\cite{imagesculpting}, yet it provides the only explicit appearance reference for newly revealed surfaces~\cite{geoedit}. A state-conditioned editor should therefore follow a spatial rule in which visible appearance follows the photograph, unseen appearance follows the state, and shadows and reflections remain consistent with the edited geometry. Standard paired training~\cite{neuralassets,blenderfusion,threedfixup} does not explicitly teach this rule, since the photograph and state typically describe the same object and provide few controlled conflicts. Moreover, the target state is aligned with the target view, whereas the photograph must be matched across viewpoints, encouraging the editor to copy inaccurate state textures onto visible surfaces. On the other hand, unseen surfaces can often be inferred from the photograph or generative priors, so the editor may not follow the state consistently across edits. Existing 3D-guided editors either refine rendered proxies at inference~\cite{geoedit,imagesculpting} or learn from paired data~\cite{blenderfusion,threedfixup}, without explicitly resolving which source to trust.

To address this, we propose \textbf{counterfactual state supervision} (\cref{fig:overview}b), which constructs photo--state pairs with controlled conflicts and known targets for flow-matching training. We consider three cases: (1) repainting surfaces hidden from the photograph, (2) adding geometry to surfaces hidden in the source photograph, with shadows and reflections re-rendered, and (3) replacing state textures on photograph-visible surfaces with image-to-3D reconstructed textures while keeping the photograph and target unchanged. The first two teach the editor to follow the state where the photograph provides no evidence, while the third teaches it to preserve visible appearance despite discrepancies with the state.

Learning which source to trust, however, does not guarantee that this preference persists throughout sampling. Newly revealed surfaces and their shadows or reflections may be lost, while visible appearance may drift toward the reconstructed state. By switching the state or photograph at different sampling steps, we observe that the editor becomes less responsive to condition changes as sampling progresses. This highlights a limitation of standard flow-matching training, which learns from noisy ground-truth targets without explicitly training the model to recover from trajectories drifting toward condition-inconsistent outputs.

We therefore propose \textbf{symmetric counterfactual correction} (\cref{fig:overview}c), built on two types of counterfactual twins. State-side twins share the photograph but differ in state and target through hidden-surface repainting or adding geometry to occluded surfaces, while photo-side twins share the state but differ in photograph and target. Using these twins, we perturb each target latent toward its paired counterpart on the changed tokens and train the model to recover the target specified by its current condition. These perturbations train recovery from condition-inconsistent deviations, reinforcing reliance on the state for unseen content and on the photograph for visible appearance through state-side and photo-side correction, respectively. Both strategies operate only during training, requiring no architectural changes or additional inference steps.

In summary, our contributions are:
\begin{itemize}

\item We introduce counterfactual state supervision to teach the editor to follow the photograph for visible appearance and the 3D state for unseen content through three types of controlled photo--state conflicts.
\item We propose symmetric counterfactual correction, using state-side and photo-side twins with directed latent perturbations to train recovery from condition-inconsistent sampling drift without modifying inference.
\item We construct a synthetic training dataset spanning 254 objects and 3,308 scenes, with counterfactual variations of 3D states and photographs to support both training strategies.
\item We demonstrate strong spatial editing performance and content preservation on public benchmarks, while controlled ablations and condition-switching analyses show improved source selection and sustained responsiveness to conditioning during sampling.

\end{itemize}

\section{Related Work}
\label{sec:related}

\noindent\textbf{3D-Aware Object Editing.}
Existing approaches to 3D-aware object editing can be broadly grouped into three categories. Prior-driven methods infer unseen content from learned image or motion priors, including novel-view synthesis~\cite{zero123}, language-guided 3D-aware editing~\cite{object3dit}, and video-based object manipulation~\cite{magicfixup,objectmover}. Recent approaches also explore fine-grained spatial control through 3D bounding boxes~\cite{thinkinginboxes} and specialized spatial editing models~\cite{spatialedit}. However, reliance on generative priors can lead to inconsistent object appearance and geometry under large transformations. Training-free methods introduce explicit geometric guidance into pretrained diffusion models by lifting diffusion features into 3D~\cite{diffusionhandles}, sharing attention through depth-based correspondences~\cite{geodiffuser}, rendering transformed geometric proxies~\cite{geoedit,imagesculpting}, or decoupling geometric transformation from generation~\cite{freefine}. Yet, geometric guidance alone does not resolve how an editor should balance photograph appearance against imperfect 3D guidance. Learned 3D-conditioned editors instead incorporate object poses~\cite{neuralassets} or rendered 3D proxies~\cite{blenderfusion,threedfixup} as conditions during training. However, conventional training pairs rarely expose conflicts between the source photograph and reconstructed 3D state, leaving the model without explicit guidance on which source to trust. In contrast, our method explicitly learns to preserve photograph-visible appearance while following the 3D state for unseen content.

\noindent\textbf{Counterfactual Supervision and Sampling Correction.}
Counterfactual augmentation isolates selected factors to encourage learning from task-relevant evidence~\cite{kaushik2020cad}. In image editing, ObjectDrop~\cite{objectdrop} exploits photographs captured before and after object removal to learn associated shadows and reflections, while conditional generation and editing methods~\cite{brooks2023ip2p,controlnet} typically learn from condition--target pairs without explicitly addressing conflicts between photographs and reconstructed 3D states. Meanwhile, standard diffusion and flow-matching training constructs noisy samples from ground-truth targets, whereas inference follows model-generated trajectories that may drift toward condition-inconsistent outputs. Prior approaches address this discrepancy through training on model-visited states~\cite{bengio2015scheduled,ross2011dagger}, simulating prediction errors with input perturbations~\cite{ning2023inputpert}, adjusting the sampling process~\cite{ning2024exposure}, or using model-generated rollouts~\cite{selfforcing}. While these approaches address training--inference discrepancies and accumulated errors, they do not explicitly train recovery from drift toward condition-inconsistent outputs. Our counterfactual state supervision teaches the editor which conditioning source to trust, while symmetric counterfactual correction uses directed latent perturbations between twin targets to train recovery from condition-inconsistent drift.
\section{Method}
\label{sec:method}

Given a photograph $I$, a registered textured 3D asset, and a geometric edit operation, including rotation, translation, or rescaling, our goal is to generate the edited image \(Y\) while preserving the background and updating shadows and reflections. We render the transformed asset as a 3D state \(S=(S^c,S^n)\), consisting of colour and camera-space normal images. The editor follows a spatial rule: visible appearance follows the photograph, unseen content follows the state, and shadows and reflections remain consistent with the edited geometry. Our method comprises a state-conditioned editor (\cref{sec:editor}), counterfactual state supervision (\cref{sec:cf}), and symmetric counterfactual correction (\cref{sec:correction}).

\subsection{State-conditioned Editor}
\label{sec:editor}

\noindent\textbf{Backbone and inputs.}
We build on FLUX.1 Kontext~\cite{kontext}, freezing the backbone and training LoRA adapters~\cite{hu2022lora}. The model denoises the target latent \(z\) conditioned on \(C=(I,S^c,S^n,P_0,M)\), where \(P_0\) is the source-object crop and \(M\) is its mask. We use a fixed prompt for all edits: ``Move the object to the target pose shown in the reference and update its shadows and reflections; keep everything else unchanged.'' The state renderings are aligned with the target image, allowing each target token to directly access its corresponding 3D-state information.

\noindent\textbf{Flow matching.}
Given a noise level \(\sigma\in(0,1)\) and Gaussian noise \(\epsilon\sim\mathcal N(0,I)\), we construct \(x_\sigma=(1-\sigma)z+\sigma\epsilon\) and train the model to predict the velocity \(\epsilon-z\) using a flow-matching objective:
\begin{equation}
\mathcal{L}_{\text{FM}} = \mathbb{E}\left[\|v_{\theta}(x_{\sigma}, \sigma; C) - (\epsilon - z)\|_2^2\right].
\end{equation}

\noindent\textbf{Cross-view correspondence.}
Unlike the state, the source photograph is not aligned with the target view. Using registered geometry, we estimate the source location \(u_j\) and visibility of each target token \(j\). For each target token \(j\) with a reliable visible correspondence, the source-photo tokens around \(u_j\) are assigned the rotary position~\cite{rope} of token \(j\), enabling cross-view appearance retrieval without moving image pixels. At these locations, state-colour tokens receive randomly displaced positional encodings, while state-normal tokens retain target-aligned encodings.

\subsection{Counterfactual State Supervision}
\label{sec:cf}

Ordinary paired training does not explicitly teach which source to trust when the photograph and state disagree. Since both describe the same object, the model can reproduce the target by relying on the spatially aligned state rather than learning to preserve photograph-visible appearance. To address this, we introduce counterfactual state supervision, which constructs counterfactual photo--state pairs with known targets.

Starting from a rendered pair $(I,S)\mapsto Y$, we construct a counterfactual 3D state $S'=(S'^c,S'^n)$ by modifying surface appearance or geometry. We retain the auxiliary rendering passes for albedo $A$, diffuse lighting $L$, and non-diffuse radiance $Q$, approximating the rendered appearance as $\mathrm{srgb}(L \odot A + Q)$. We consider three categories:

\noindent\textbf{Hidden-surface repainting.}
We modify the albedo of photograph-invisible surfaces while keeping the photograph unchanged. With a new albedo \(A'\), the target is updated under fixed light transport:
\begin{equation}
Y^{\prime} = Y + \mathrm{srgb}(L \odot A^{\prime} + Q) - \mathrm{srgb}(L \odot A + Q),
\end{equation}
where $\mathrm{srgb}(\cdot)$ converts linear radiance to sRGB. Since the correction vanishes wherever $A'=A$, unaffected pixels are preserved without re-rendering noise. The state $S'$ is updated consistently, requiring the editor to follow the state for newly revealed appearance.

\noindent\textbf{Hidden geometry.}
We attach additional geometry to source-occluded surfaces and re-render the target with its shadows and reflections. We retain placements whose source-view rendering remains nearly unchanged, so the new geometry and its scene effects must be inferred from the state rather than the source photograph, where the added geometry remains hidden.

\noindent\textbf{Visible-surface texture replacement.}
We replace state textures on photograph-visible surfaces with Hunyuan3D~2.1 reconstructions~\cite{hunyuan3d21}, while keeping the photograph and target unchanged. The unchanged target supervises the editor to preserve photograph-visible appearance despite reconstruction errors.

Together, the first two categories supervise reliance on the state for unseen content, while the third supervises reliance on the photograph for visible appearance.

\subsection{Symmetric Counterfactual Correction}
\label{sec:correction}

Counterfactual supervision teaches the model which source to trust but does not train recovery from condition-inconsistent sampling drift. We address this through symmetric counterfactual correction using pairs of counterfactual twins.

\begin{figure*}[t]
    \centering
    \includegraphics[width=\textwidth,height=7cm]{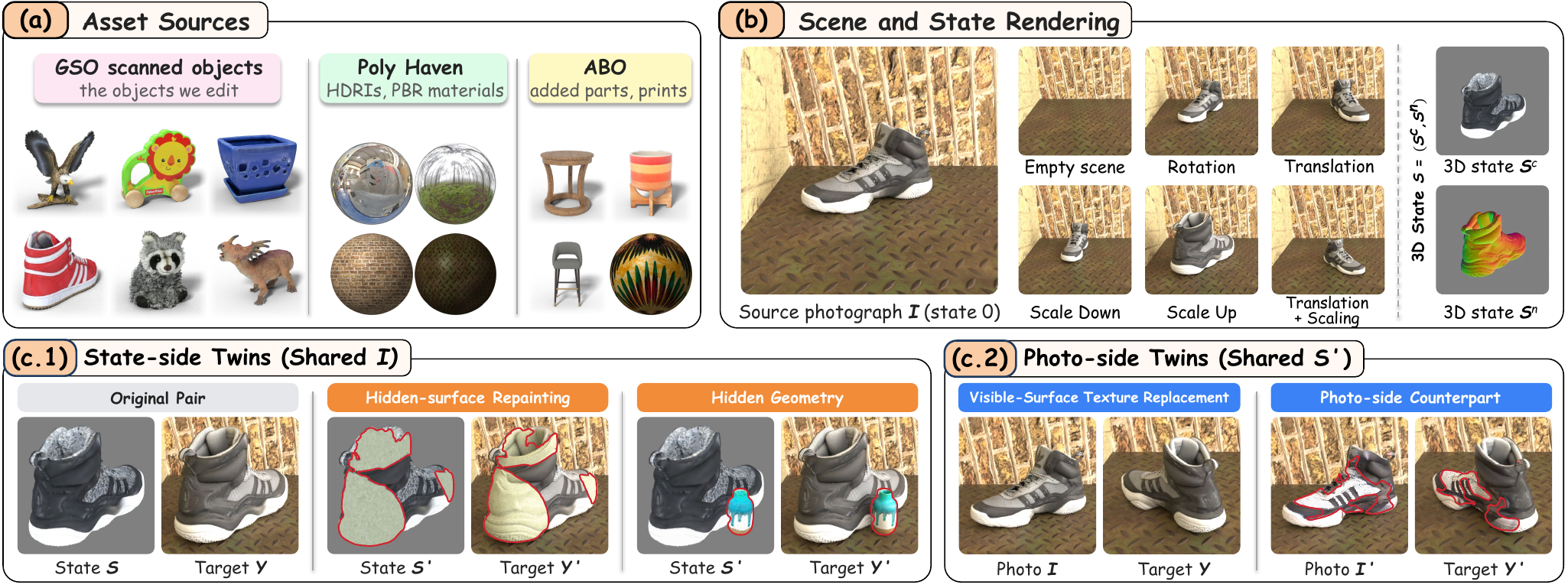}
    \caption{\textbf{Dataset construction.}
    (a)~Sources of scanned objects, materials, lighting, and additional parts.
    (b)~Each scene contains 12 object states under fixed camera and lighting, with colour and normal renderings $S=(S^c,S^n)$.
    (c)~State-side twins share the photograph but differ in state and target, while photo-side twins share the state but differ in photograph and target. Only state-colour renderings are shown, with red contours marking modified regions.
    }
    \label{fig:dataset}
    \vspace{-10pt}
\end{figure*}

\noindent\textbf{Counterfactual twins.}
We construct two types of twin pairs. State-side twins share the photograph but differ in the state and target, pairing $(I,S)\mapsto z$ with $(I,S')\mapsto z'$ through hidden-surface repainting or hidden geometry. Photo-side twins instead share the state $S'$ but differ in photograph and target, pairing $(I,S')\mapsto z$ with $(I',S')\mapsto z'$. \(I'\) and \(z'\) are constructed by transferring reconstructed textures to the corresponding visible regions of the photograph and target. A binary mask $m$ identifies the tokens that differ between the two targets.

\noindent\textbf{Recovery from counterfactual drift.}
With $\alpha \sim \mathcal{U}(0,1)$, we construct two perturbed latents symmetrically:

\begin{equation}
\begin{aligned}
\tilde z  &= z + \alpha m \odot (z' - z),\\
\tilde z' &= z' - \alpha m \odot (z' - z).
\end{aligned}
\end{equation}

The perturbations move each target toward the other target only on the changed tokens. We then add the same Gaussian noise \(\epsilon\) to both:

\begin{equation}
\begin{aligned}
x_\sigma&=(1-\sigma)\tilde z+\sigma\epsilon,\\
x'_\sigma&=(1-\sigma)\tilde z'+\sigma\epsilon.
\end{aligned}
\end{equation}

Despite these perturbations, each twin is supervised to recover its own target under its corresponding condition. For the first twin, we require the clean estimate \(x_\sigma-\sigma v^{*}\) to equal \(z\), yielding
\begin{equation}
\begin{aligned}
v^{*}
=
\frac{x_\sigma-z}{\sigma}
=
\underbrace{(\epsilon-z)}_{\text{flow matching}}
+
\underbrace{\frac{1-\sigma}{\sigma}\alpha m\odot(z'-z)}_{\text{counterfactual correction}}.
\end{aligned}
\end{equation}
The first term is the standard flow-matching target, while the second corrects the drift toward $z'$ on changed tokens. Its magnitude increases as $\sigma$ decreases, requiring stronger correction at lower noise levels.

Symmetrically, the second twin is trained to recover $z'$ from its perturbed latent. Finally, we optimize both directions using a recovery loss:
\begin{equation}
\begin{aligned}
\mathcal{L}_{\text{corr}} &= \mathbb{E}\left[\|v_{\theta}(x_{\sigma}, \sigma; C) - v^{*}\|_2^2\right.\\
&\quad\left.{}+ \|v_{\theta}(x_{\sigma}^{\prime}, \sigma; C^{\prime}) - v^{\prime*}\|_2^2\right],
\end{aligned}
\end{equation}
where $C$ and $C^{\prime}$ are the respective conditioning inputs, and $v^{\prime*} = (\epsilon - z^{\prime}) - \frac{1-\sigma}{\sigma}\alpha m \odot (z^{\prime} - z)$ is the symmetric target velocity for the second twin.

In this way, state-side correction teaches recovery toward state-consistent targets for unseen content, while photo-side correction teaches recovery toward photograph-consistent targets for visible appearance. Both are applied only during training, leaving the network architecture and inference procedure unchanged.

\begin{table*}[t]
  \centering
  \small
  \caption{\textbf{Quantitative comparison on public benchmarks.}
  Results on OBJect and SpatialEdit-Bench.
  Best and second-best results are highlighted in bold and underlined, respectively.}
  \label{tab:public_benchmarks}
  \setlength{\tabcolsep}{3pt}
  \renewcommand{\arraystretch}{0.9}
  \resizebox{\textwidth}{!}{%
  \begin{tabular}{@{}lcccccc@{\hspace{10pt}}cccccc@{}}
    \toprule
    \multirow[c]{2}{*}[-0.9ex]{Method}
    & \multicolumn{6}{c}{\textbf{OBJect}}
    & \multicolumn{6}{c}{\textbf{SpatialEdit-Bench}} \\
    \cmidrule(lr){2-7}\cmidrule(l){8-13}
    & LPIPS$^{\times 10^3}_{\rm bg}\downarrow$
    & DINO$_{\rm obj}\uparrow$
    & $S_{\rm view}\uparrow$
    & RS$\uparrow$
    & $S_{\rm oc}\uparrow$
    & MD$\downarrow$
    & LPIPS$^{\times 10^3}_{\rm bg}\downarrow$
    & DINO$_{\rm obj}\uparrow$
    & $S_{\rm view}\uparrow$
    & RS$\uparrow$
    & $S_{\rm oc}\uparrow$
    & MD$\downarrow$ \\
    \midrule
    SpatialEdit
    & 54.70 & 0.791 & 0.190 & 0.160 & \underline{0.567} & \underline{34.33}
    & \underline{31.84} & 0.832 & \textbf{0.889}
    & \textbf{0.826} & \underline{0.668} & \underline{25.93} \\

    TiB
    & 121.9 & \underline{0.871} & \underline{0.460}
    & \underline{0.435} & 0.447 & 41.5
    & 117.8 & \underline{0.840} & 0.514 & 0.468 & 0.447 & 111.1 \\

    GeoDiffuser
    & 76.70 & 0.776 & 0.370 & 0.298 & 0.358 & 63.59
    & 69.74 & 0.811 & 0.498 & 0.375 & 0.418 & 75.36 \\

    FreeFine
    & 87.20 & 0.673 & 0.240 & 0.161 & 0.257 & 82.80
    & 93.58 & 0.834 & 0.463 & 0.317 & 0.551 & 31.74 \\

    DiffHandles
    & 322.0 & 0.615 & 0.280 & 0.153 & 0.100 & 94.27
    & 322.3 & 0.643 & 0.232 & 0.130 & 0.095 & 72.12 \\

    FLUX.1 Kontext
    & \underline{45.04} & 0.746 & 0.200 & 0.188 & 0.203 & 172.2
    & 73.43 & 0.785 & 0.508 & 0.466 & 0.262 & 172.9 \\

    Qwen-Image-Edit
    & 158.8 & 0.716 & 0.060 & 0.051 & 0.415 & 151.9
    & 60.19 & 0.799 & 0.697 & 0.652 & 0.487 & 127.4 \\

    \midrule
    \textbf{Ours}
    & \textbf{17.59} & \textbf{0.878}
    & \textbf{0.580} & \textbf{0.527}
    & \textbf{0.579} & \textbf{28.48}
    & \textbf{27.61} & \textbf{0.851}
    & \underline{0.867} & \underline{0.768}
    & \textbf{0.688} & \textbf{24.17} \\
    \bottomrule
  \end{tabular}}
  \vspace{-10pt}
\end{table*}

\subsection{Training and Inference}
\label{sec:impl}

\noindent\textbf{Training.} We adopt a two-stage procedure. Stage~1 trains the editor with counterfactual state supervision using $\mathcal L_{\mathrm{FM}}$. Stage~2 initializes from Stage~1 and jointly optimizes $\mathcal L_{\mathrm{FM}}$ and $\mathcal L_{\mathrm{corr}}$ using both state-side and photo-side twins.

\noindent\textbf{Inference.}
Given a photograph, SAM~\cite{sam} segments the object, and Hunyuan3D~2.1~\cite{hunyuan3d21} reconstructs a textured 3D asset from its crop.
We register the asset to the source photograph using the object mask and depth estimates from MoGe-2~\cite{moge2}. The registered asset is transformed according to the desired rotation, translation, or scaling and rendered as $S^c$
and $S^n$. The editor then generates the final image without additional test-time optimization.

\section{Experiments}
\label{sec:experiments}

\begin{figure*}[t]
    \centering    \includegraphics[width=\textwidth,height=9cm]{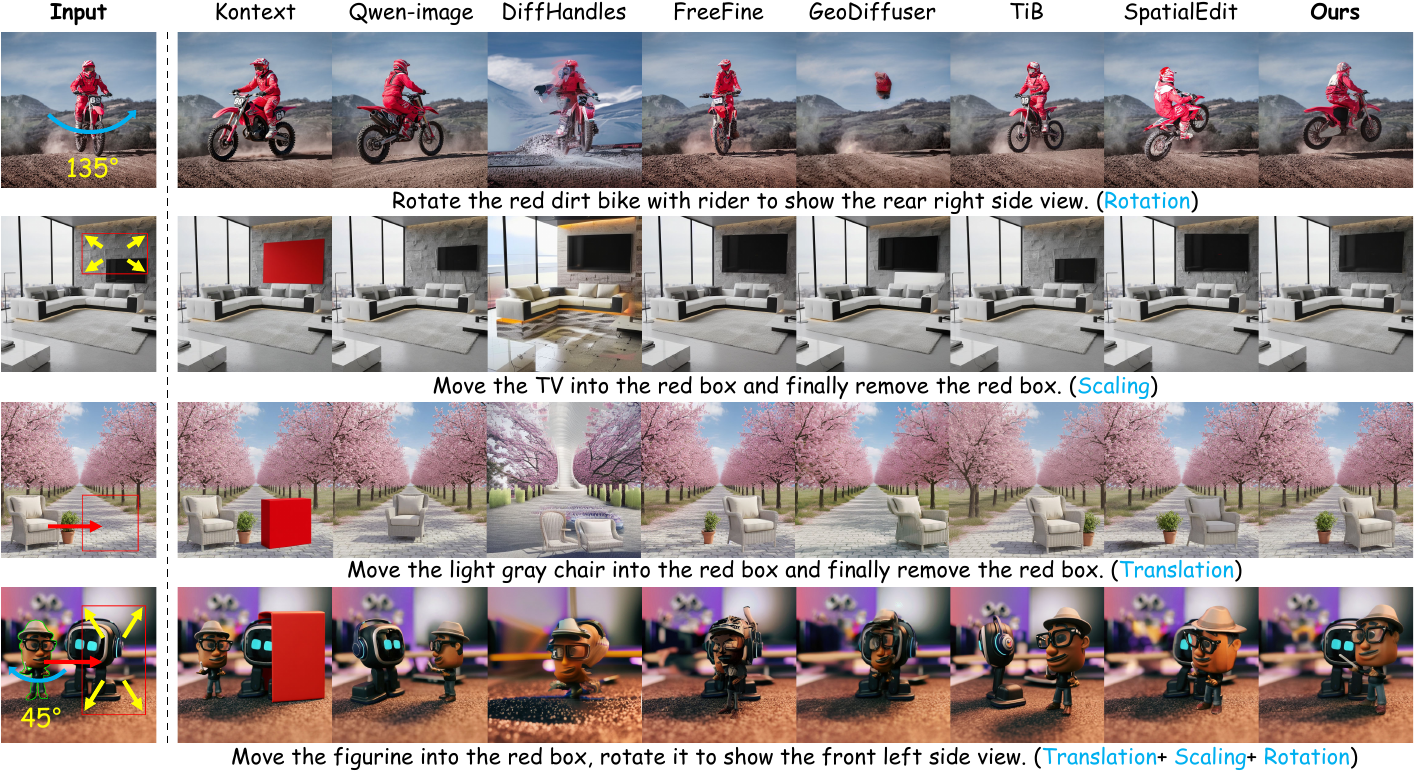}
    \caption{\textbf{Qualitative comparison with existing spatial editing methods.}
    From top to bottom, we show rotation, scaling, translation, and their combination.
    Red boxes indicate target regions, while arrows illustrate the requested transformations.
    }
    \label{fig:qual_comparison}
    \vspace{-10pt}
\end{figure*}

\subsection{Datasets}
\label{sec:datasets}

\noindent\textbf{Training Data.} We construct 3,308 training scenes using 254 scanned objects from GSO~\cite{gso} across twelve scene types. Poly Haven PBR materials and HDRIs provide surface appearance and lighting, respectively. Each scene is rendered under fixed camera and lighting conditions with 12 object states generated through rotation, translation, and scaling (\cref{fig:dataset}a,b). State~0 serves as the source photograph, yielding 36,388 editing pairs with the other 11 states.

For counterfactual supervision (\cref{fig:dataset}c), we construct 5,414 \emph{hidden-surface repainting} samples using Poly Haven textures or ABO~\cite{abo} product prints, 2,624 \emph{hidden-geometry} samples with ABO 3D models attached to occluded surfaces, and 2,712 \emph{visible-surface texture replacement} samples using Hunyuan3D~2.1~\cite{hunyuan3d21} textures. We additionally construct 2,128 \emph{photo-side counterparts} by transferring reconstructed textures to the photograph and target. Repainted and photo-side images are composited using Eq.~(2) in non-reflective scenes, while hidden-geometry samples are re-rendered. The resulting dataset contains approximately 172K training pairs, all using State~0 as the source photograph.

\noindent\textbf{Evaluation Set.}
We evaluate on 24 held-out objects across 240 scenes, comprising 1,920 cases in four settings: hidden-surface repainting (480), hidden geometry (480), visible-surface appearance preservation (480), and standard geometric editing (480). These assess unseen-content generation, photograph-visible appearance preservation, and editing quality.

\noindent\textbf{Public Benchmarks.}
We evaluate on two public benchmarks.
OBJect~\cite{object3dit} provides ground-truth edited images of Objaverse objects in synthetic scenes. We sample 200 test cases for evaluation, including 100 rotations and 100 translations.
For SpatialEdit-Bench~\cite{spatialedit}, we evaluate 764 edits from 103 source images, including 488 rotations to eight canonical views and 276 translations to target boxes, without ground-truth edited images.

\subsection{Experimental Setup}
\label{sec:experimental_setup}

\noindent\textbf{Implementation Details.}
We train rank-16 LoRA adapters on the query, key, value, and output projections of FLUX.1 Kontext~\cite{kontext} at $512\times512$ resolution, using AdamW with a learning rate and weight decay of $10^{-4}$. Stage~1 optimizes $\mathcal L_{\mathrm{FM}}$ on standard and counterfactual editing pairs for 12,000 iterations, excluding photo-side counterparts. Stage~2 resumes from the first-stage model and optimizer states for another 6,000 iterations, jointly optimizing $\mathcal L_{\mathrm{FM}}+\mathcal L_{\mathrm{corr}}$. The correction loss uses one state-side and one photo-side twin pair when available in the sampled scene (\cref{sec:correction}). Training uses two NVIDIA A800 (80\,GB) GPUs. At inference, we use a guidance scale of 2.5. Given a reconstructed Hunyuan3D~2.1~\cite{hunyuan3d21} asset, each edit takes approximately 19\,s on a single A800.

\noindent\textbf{Comparison Methods.}
We compare against seven methods on both benchmarks. Instruction-based methods include FLUX.1 Kontext~\cite{kontext}, Qwen-Image-Edit-2511~\cite{qwenimage}, and SpatialEdit-16B~\cite{spatialedit}. Geometric editing methods include GeoDiffuser~\cite{geodiffuser}, Diffusion Handles~\cite{diffusionhandles}, FreeFine~\cite{freefine}, and Thinking in Boxes~\cite{thinkinginboxes}. We use official implementations and pretrained weights with method-specific inputs.





\noindent\textbf{Evaluation Metrics.} We evaluate edit accuracy, content preservation, and overall score using six metrics across both benchmarks.
For edit accuracy, the viewpoint score ($S_{\mathrm{view}}$) measures rotation correctness through binary VLM judgments, using official SpatialEdit-Bench questions or ground-truth reference images for OBJect. Mean Distance (MD)~\cite{freefine} measures translation error between DIFT-matched SIFT keypoints and their expected target locations~\cite{dift}.

For content preservation, $\mathrm{DINO}_{\mathrm{obj}}$ measures DINOv3 feature similarity between source and edited objects~\cite{dinov3}, while $\mathrm{LPIPS}_{\mathrm{bg}}$ measures perceptual changes outside the source and target boxes~\cite{lpips}. For translation, we also report the VLM-judged object consistency score $S_{\mathrm{oc}}$~\cite{spatialedit}, which rates whether the moved object's identity and the background are preserved.

For the overall score, we report the rotation score $\mathrm{RS}=\sqrt{S_{\mathrm{view}}S_{\mathrm{cons}}}$ following SpatialEdit-Bench~\cite{spatialedit}, where $S_{\mathrm{cons}}$ is the VLM-judged consistency of object identity and background. The geometric mean rewards only edits that are both correct and consistent. Both $S_{\mathrm{cons}}$ and $S_{\mathrm{oc}}$ are normalized to $[0,1]$. We use Qwen3.8-27B~\cite{qwen38} for all VLM judgments.

\noindent\textbf{Quantitative Results.}
As shown in \cref{tab:public_benchmarks}, our method achieves the best results across all six metrics on OBJect, including $S_{\mathrm{view}}$ (0.580) and RS (0.527) for rotation, and MD (28.48) for translation. It also achieves the highest DINO$_{\mathrm{obj}}$ (0.878) and lowest LPIPS$_{\mathrm{bg}}$ (17.59). On SpatialEdit-Bench, our method achieves the best results in four of six metrics, including DINO$_{\mathrm{obj}}$ (0.851), LPIPS$_{\mathrm{bg}}$ (27.61), MD (24.17), and $S_{\mathrm{oc}}$ (0.688). For rotation, it ranks second in $S_{\mathrm{view}}$ (0.867) and RS (0.768), behind SpatialEdit. Notably, SpatialEdit is fine-tuned on SpatialEdit-500k, whose canonical-view rotations and box-guided transformations align with SpatialEdit-Bench~\cite{spatialedit}, whereas OBJect evaluates arbitrary-angle rotations~\cite{object3dit}. SpatialEdit performs considerably worse than our method on OBJect, highlighting its limitations in handling arbitrary-angle rotations.

\noindent\textbf{Qualitative Results.}
\Cref{fig:qual_comparison} compares our method with existing approaches across rotation, scaling, translation, and their combination. 
Instruction-based methods often struggle with precise spatial control, as FLUX.1 Kontext reproduces the red target box and Qwen-Image-Edit alters unrelated content. Geometry-controlled methods also exhibit limitations: Diffusion Handles and GeoDiffuser distort or lose the dirt bike under large rotations, while FreeFine and TiB fail to achieve the requested orientation. SpatialEdit follows spatial instructions more reliably but alters the appearance of the rider and figurine. In contrast, our method successfully performs all four edits, preserving object appearance and background content while updating shadows and occlusions.

\subsection{Ablation Study}
\label{sec:ablation}

We ablate the contributions of counterfactual state supervision and symmetric counterfactual correction on the held-out evaluation set (\cref{tab:ablation}). Paired training uses the same standard rendering pairs and training schedule but excludes counterfactual samples. CF supervision (12k) corresponds to Stage~1, while CF supervision (18k) continues training without correction. All remaining variants resume from Stage~1 for 6,000 iterations, progressively adding state-side and photo-side correction. The $\alpha=0$ variants only use the corresponding twin pairs for standard flow-matching training without latent perturbation, under matched data and compute budgets.

\begin{table}[t]
  \centering
  \caption{\textbf{Ablation on the held-out evaluation set.}
  Each evaluation setting comprises 480 test cases from 24 unseen objects.
  Rows with $\alpha=0$ use twin pairs without latent perturbation.}
  \label{tab:ablation}
  \setlength{\tabcolsep}{3pt}
  \renewcommand{\arraystretch}{0.9}
  \resizebox{\linewidth}{!}{%
  \begin{tabular}{@{}lccccccc@{}}
    \toprule
    \multirow[c]{2}{*}[-0.9ex]{\textbf{Variant}}
      & \multicolumn{3}{c}{\textbf{State adherence \%}}
      & \multicolumn{4}{c}{\textbf{Edit fidelity}} \\
    \cmidrule(lr){2-4}\cmidrule(l){5-8}
      & HSA$\uparrow$ & HGA$\uparrow$ & VSA$\downarrow$
      & PSNR$_{\rm obj}\!\uparrow$
      & LPIPS$^{\times 10^3}_{\rm obj}\!\downarrow$
      & PSNR$_{\rm bg}\!\uparrow$
      & LPIPS$^{\times 10^3}_{\rm bg}\!\downarrow$ \\
    \midrule
    Paired training
      & 41.51 & 46.50 & 22.02 & 17.65 & 155.5 & 33.69 & 25.39 \\
    CF supervision (12k)
      & 57.57 & 48.29 & 16.64 & 17.56 & 162.7 & 33.39 & 26.75 \\
    CF supervision (18k)
      & 57.10 & 48.44 & 16.25 & 17.60 & 161.6 & 33.36 & 26.60 \\
    \midrule
    State twins ($\alpha=0$)
      & 61.34 & 49.02 & 17.15 & 17.72 & 156.4 & 33.52 & 25.57 \\
    State correction
      & \textbf{63.98} & 50.58 & 19.00 & 17.84 & 152.8 & 33.61 & 25.29 \\
    State correction + photo twins ($\alpha=0$)
      & 63.34 & \textbf{50.93} & 15.23 & 17.84 & 153.1
      & 33.72 & 25.25 \\
    \textbf{State + photo correction (Ours)}
      & 63.62 & 50.71 & \textbf{11.30} & \textbf{17.92}
      & \textbf{151.5} & \textbf{33.72} & \textbf{25.12} \\
    \bottomrule
  \end{tabular}}
  \par\vspace{7pt}
\begin{minipage}{\columnwidth}
    \captionsetup{type=figure}\centering
    \includegraphics[width=\linewidth]{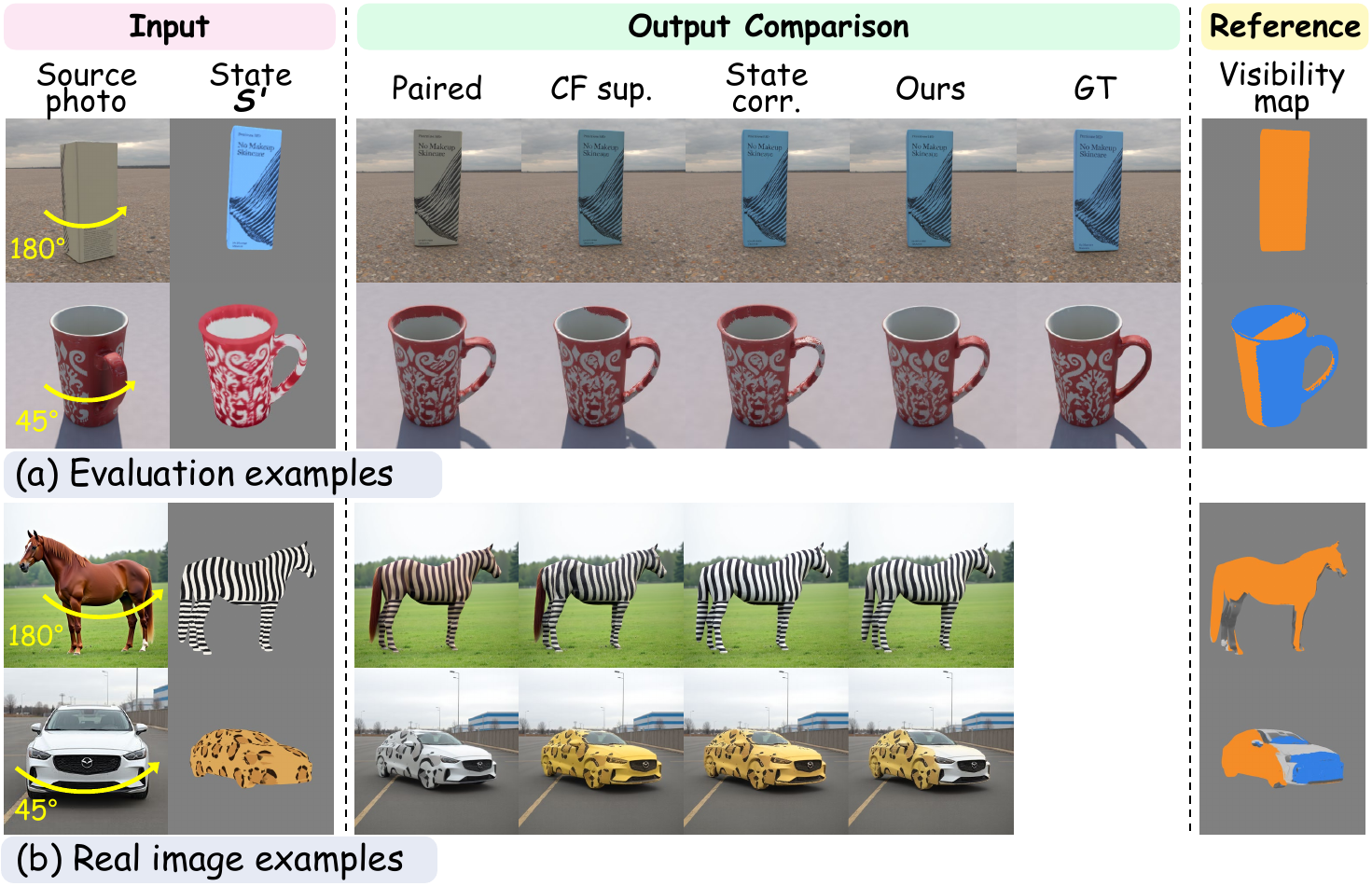}
    \caption{\textbf{Qualitative ablation of counterfactual supervision and correction.}
    The top rows in (a,b) illustrate hidden-surface adherence (HSA), while the bottom rows highlight visible-surface preservation (VSA).
    Orange and blue indicate regions that should follow the state and photograph, respectively. Ground truth is available only for (a).}
    \label{fig:qual_ablation}
    
\end{minipage}
\end{table}

\noindent\textbf{Evaluation metrics on held-out scenes.}
We measure state adherence (SA) to quantify how faithfully the editor follows controlled changes to the 3D state:
\begin{equation}
  \mathrm{SA} = \frac{\langle \hat{Y}' - \hat{Y},\, \Delta \rangle_{\Omega}}{\langle \Delta,\, \Delta \rangle_{\Omega}},
  \label{eq:adherence}
\end{equation}
where $\hat{Y}$ and $\hat{Y}'$ are the outputs generated under the original and modified states, $\Delta$ denotes the reference difference, and $\Omega$ is the evaluation region in the target view. Perfect reproduction yields 100\% SA, while no response aligned with $\Delta$ yields 0\%. We evaluate three settings:
(1) Hidden-surface adherence (HSA) evaluates repainting photograph-invisible surfaces with scanned textures from unseen objects.
(2) Hidden-geometry adherence (HGA) evaluates adding unseen scanned objects to occluded surfaces. For both, $\Delta=Y'-Y$ is the ground-truth target-view difference, and $\Omega$ covers the modified surfaces or added geometry.
(3) Visible-surface adherence (VSA) replaces the ground-truth state texture with the aligned Hunyuan3D~2.1 reconstruction used at inference. Here, $\Delta=S'^c-S^c$ is the state-colour discrepancy, and $\Omega$ covers photograph-visible surfaces affected by reconstruction errors. Since the photograph provides the true appearance, lower VSA indicates less reconstruction error leakage. Thus, high HSA/HGA indicate reliance on the state for unseen content, while low VSA indicates preservation of photograph-visible appearance. For edit fidelity, we report PSNR and LPIPS against ground truth on the target object and background.

\begin{table}[t]
  \centering
  \caption{\textbf{End-to-end editing.}
  Region-wise PSNR and LPIPS on SpatialEdit-Bench (488 rotations) and OBJect (100 rotations).}
  \label{tab:e2e}
  \setlength{\tabcolsep}{2.5pt}
  \resizebox{\linewidth}{!}{%
  \begin{tabular}{@{}lcccccccc@{}}
    \toprule
    \multirow[c]{3}{*}[-0.9ex]{\textbf{Variant}}
    & \multicolumn{4}{c}{\textbf{SpatialEdit-Bench}}
    & \multicolumn{4}{c}{\textbf{OBJect}} \\
    \cmidrule(lr){2-5}\cmidrule(l){6-9}
    & \multicolumn{2}{c}{Visible}
    & \multicolumn{2}{c}{Hidden}
    & \multicolumn{2}{c}{Visible}
    & \multicolumn{2}{c}{Hidden} \\
    \cmidrule(lr){2-3}\cmidrule(lr){4-5}
    \cmidrule(lr){6-7}\cmidrule(l){8-9}
    & PSNR$\uparrow$
    & LPIPS$^{\times 10^3}\!\downarrow$
    & PSNR$\uparrow$
    & LPIPS$^{\times 10^3}\!\downarrow$
    & PSNR$\uparrow$
    & LPIPS$^{\times 10^3}\!\downarrow$
    & PSNR$\uparrow$
    & LPIPS$^{\times 10^3}\!\downarrow$ \\
    \midrule
    Paired training
    & 14.92 & 123.71 & 13.90 & 180.62
    & 19.00 & 173.10 & 15.65 & 160.70 \\

    CF supervision (18k)
    & 14.89 & 123.87 & 13.79 & 187.25
    & 19.26 & 172.36 & 15.51 & 171.09 \\

    \textbf{Ours}
    & \textbf{15.62} & \textbf{114.84}
    & \textbf{14.53} & \textbf{173.73}
    & \textbf{20.06} & \textbf{156.87}
    & \textbf{16.53} & \textbf{156.99} \\
    \bottomrule
  \end{tabular}}
  \par\vspace{7pt}
\begin{minipage}{\columnwidth}
    \captionsetup{type=figure}\centering
    \includegraphics[width=\linewidth]{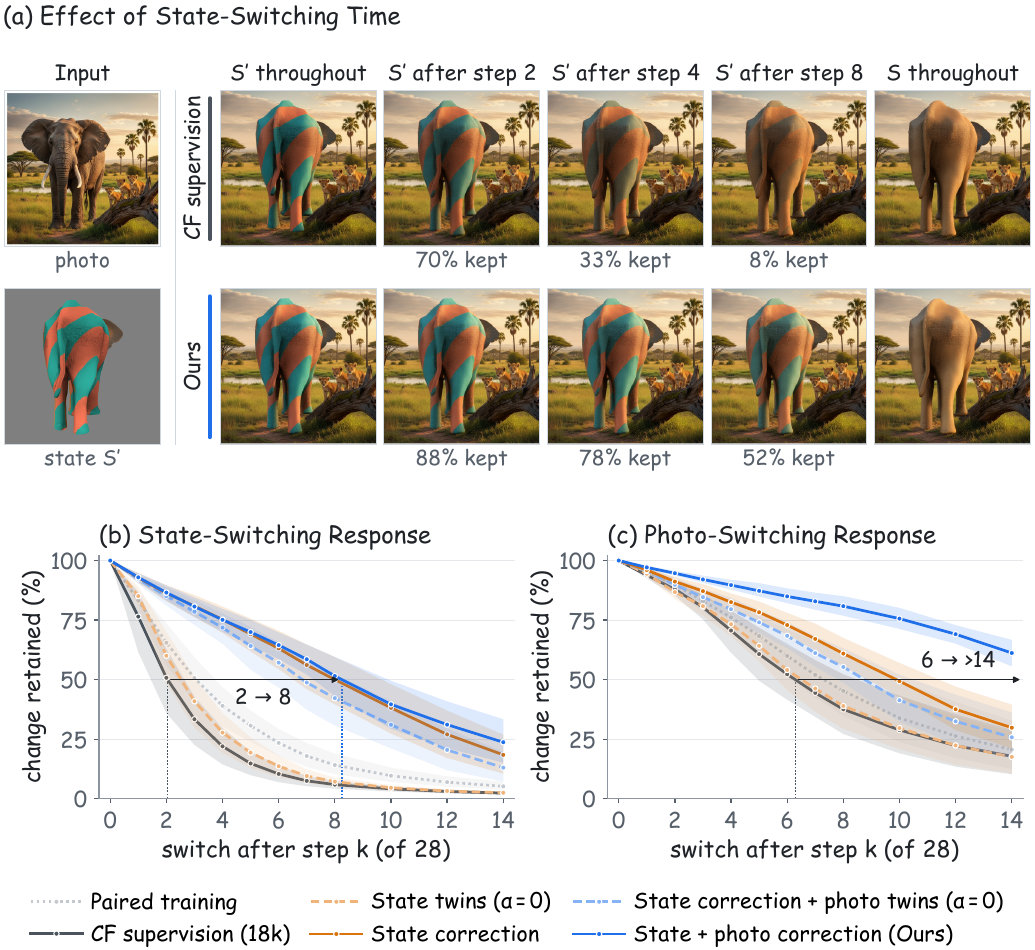}
    \caption{\textbf{Condition-switching analysis.} (a) A held-out example of switching the state at different sampling steps. (b,c) Retained change in the final output when switching the state or photograph at different sampling steps, respectively.}
    \label{fig:ablation_switch}
    
\end{minipage}
\end{table}

\noindent\textbf{Effectiveness of counterfactual state supervision.}
As shown in \cref{tab:ablation}, CF supervision (18k) increases HSA from 41.5\% to 57.1\% and reduces VSA from 22.0\% to 16.2\% compared with paired training. These improvements show that controlled photo--state conflicts teach the editor to follow the state for unseen appearance while preserving photograph-visible content. HGA also improves by 1.9 percentage points, suggesting better adherence to geometry absent from the source photograph. Qualitative examples in \cref{fig:qual_ablation} further illustrate these improvements.

\noindent\textbf{Effectiveness of symmetric counterfactual correction.}
Extending counterfactual supervision from 12k to 18k iterations yields negligible changes in hidden-content adherence. In contrast, allocating the same 6,000 iterations to symmetric counterfactual correction increases HSA from 57.6\% to 63.6\% and reduces VSA from 16.6\% to 11.3\%, with further gains in HGA. The two correction directions exhibit clear complementarity. State-side correction strengthens reliance on the state but increases visible error leakage, whereas photo-side correction suppresses this leakage while largely retaining the state-side gains, as illustrated in \cref{fig:qual_ablation}. Comparisons with the $\alpha=0$ variants further demonstrate that directed latent perturbation contributes beyond twin pairs data augmentation by training recovery from counterfactual drift. Meanwhile, the full model maintains comparable object and background edit fidelity to paired training.

\noindent\textbf{Condition-switching analysis.} We examine sampling dynamics by switching the state or photograph at different steps of the sampling process (\cref{fig:ablation_switch}). Compared with CF supervision, symmetric counterfactual correction maintains stronger responses to late condition changes. State-side correction improves responsiveness to state switches, while photo-side correction strengthens responsiveness to photograph switches, supporting their complementary roles in recovering from condition-inconsistent drift.

\noindent\textbf{End-to-end editing.}
We further evaluate our training strategies with automatically reconstructed 3D states on 488 SpatialEdit-Bench and 100 OBJect rotations. All variants share identical reconstructed assets, registration, and sampling seeds. Photograph-visible surfaces are compared with ground truth on OBJect or registration-based photograph warps on SpatialEdit-Bench, while unseen surfaces are compared with the rendered state $\Sc$. As shown in \cref{tab:e2e}, our full model achieves the best PSNR and LPIPS across both regions and benchmarks. Notably, CF supervision alone yields slightly worse hidden-surface fidelity than paired training, consistent with its early loss of state responsiveness in \cref{fig:ablation_switch}. Coarse appearance may follow the state in early steps, while finer texture details become less responsive at later steps. Symmetric correction sustains state responsiveness over more sampling steps, helping recover these details and improving fidelity on both visible and unseen surfaces.

\section{Conclusion}
\label{sec:conclusion}

We presented a 3D-aware object editing framework that addresses conflicts between source photographs and reconstructed 3D states. Counterfactual state supervision teaches the editor to follow the state for unseen content while preserving photograph-visible appearance. Symmetric counterfactual correction uses state-side and photo-side twins to train recovery from condition-inconsistent sampling drift. Both strategies operate only during training, requiring no architectural changes or additional inference steps. Experiments on OBJect and SpatialEdit-Bench demonstrate strong spatial editing performance and content preservation. 
Controlled ablations confirm the benefits of both strategies, condition-switching analyses demonstrate improved late-stage responsiveness, and end-to-end evaluations show improved fidelity with reconstructed 3D states.


\begingroup
\small
\bibliographystyle{ieeenat_fullname}
\bibliography{main}
\endgroup
\clearpage
\appendix
\twocolumn[{\centering\fontsize{14.4}{16}\selectfont\bfseries Supplementary Material\par\vspace{12pt}}]
\setcounter{topnumber}{1}
\noindent This supplementary material provides additional experimental results.
We first present additional qualitative comparisons on SpatialEdit-Bench~\cite{spatialedit} and OBJect~\cite{object3dit} (\cref{sec:supp_qual}).
We then evaluate cross-view consistency through a user study with 21 participants and a VLM judge (\cref{sec:supp_xview}).
We also examine whether the editor follows the 3D state on hidden surfaces while preserving the photographed appearance on visible surfaces in SpatialEdit-Bench (\cref{sec:supp_state_adherence}).
Finally, we discuss the limitations of our method arising from imperfect image-to-3D reconstruction (\cref{sec:supp_limitations}).

\section{Additional Qualitative Comparisons}
\label{sec:supp_qual}
\begin{figure*}[tp]
  \centering  \includegraphics[width=\textwidth,height=0.88\textheight,keepaspectratio]{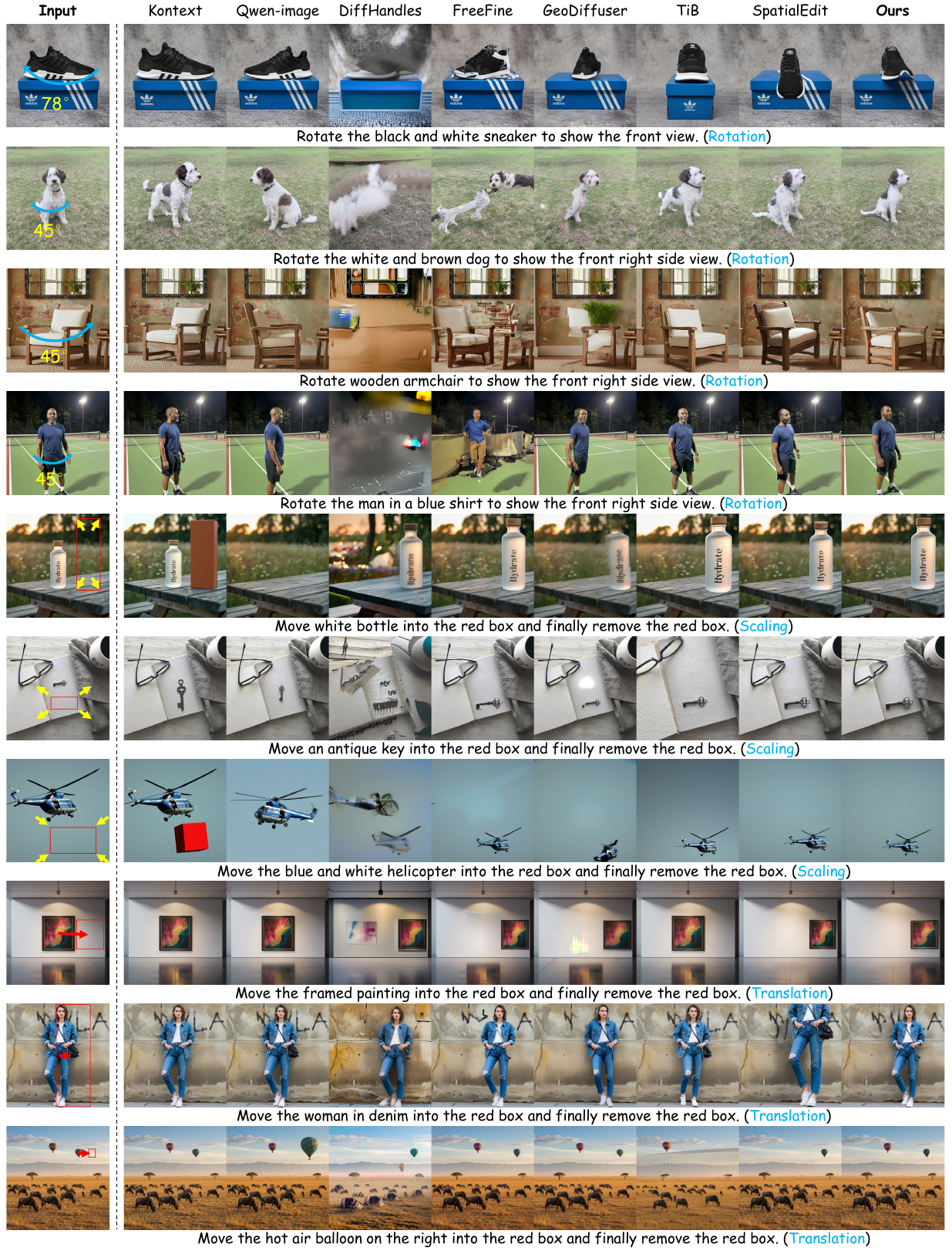}
  \caption{\textbf{Additional qualitative comparisons on SpatialEdit-Bench.}
  Rows 1--4 show rotation, rows 5--7 show scaling, and rows 8--10 show translation.
  }
  \label{fig:supp_qual_se}
\end{figure*}

\begin{figure*}[tp]
  \centering  \includegraphics[width=\textwidth,height=0.88\textheight,keepaspectratio]{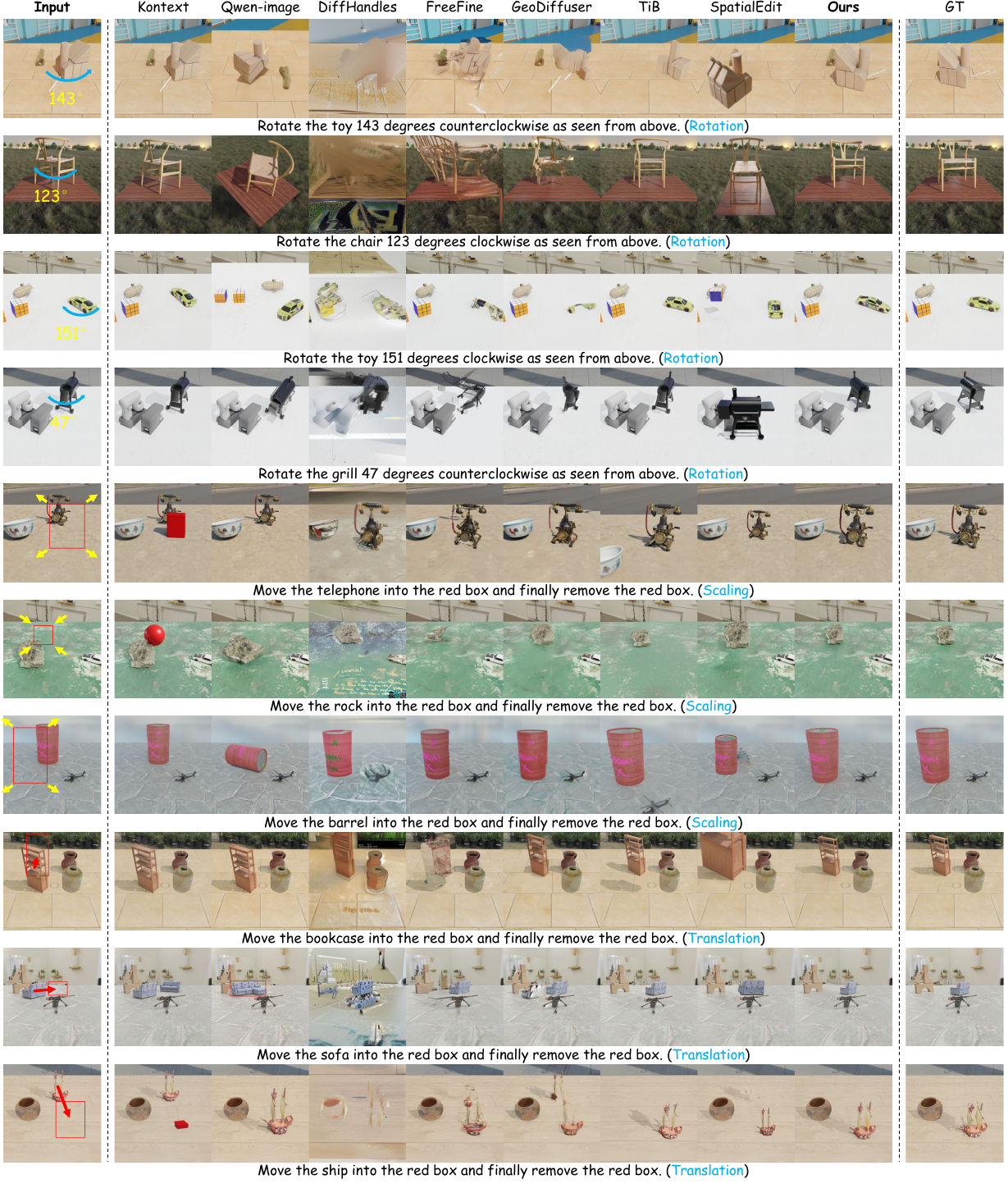}
  \caption{\textbf{Additional qualitative comparisons on OBJect.}
  Rows 1--4 show rotation, rows 5--7 show scaling, and rows 8--10 show translation.}
  \label{fig:supp_qual_obj}
\end{figure*}

\Cref{fig:supp_qual_se,fig:supp_qual_obj} provide additional qualitative comparisons on SpatialEdit-Bench and OBJect-3DIT, covering rotation, scaling, and translation.
These examples complement the results in the main paper and allow a more comprehensive visual comparison with existing methods.
Overall, they further demonstrate the effectiveness of our method in following spatial editing instructions while preserving object appearance and background content.

\section{Cross-View Consistency: User Study and VLM Judge}
\label{sec:supp_xview}
When an object is edited into different viewpoints, the results should depict the same physical object with consistent shape and appearance. Image-only editors must infer previously unseen surfaces, whereas our method conditions each edit on renderings of the same reconstructed 3D asset. We evaluate viewpoint accuracy and cross-view consistency through a two-alternative forced-choice (2AFC) study with 21 participants and a VLM judge, using the same questions and presentation format.

\noindent\textbf{Evaluation protocol.}
We use 61 source photographs from the rotation subset of SpatialEdit-Bench. Each photograph has eight rotation edits targeting views at $45^\circ$ intervals (right side, front right, front, front left, left side, rear left, rear, and rear right).
For each method, we use the same outputs as those evaluated in Table~1 of the main paper, crop them to a common window around the object, and arrange the eight views in a grid labelled by target viewpoint.
Each comparison presents the source photograph alongside two grids, A and B, showing our results and those of a baseline.

We compare against seven methods (FLUX.1 Kontext~\cite{kontext}, Qwen-Image-Edit-2511~\cite{qwenimage}, SpatialEdit~\cite{spatialedit}, TiB~\cite{thinkinginboxes}, GeoDiffuser~\cite{geodiffuser}, Diffusion Handles~\cite{diffusionhandles}, and FreeFine~\cite{freefine}) and a paired-training ablation that shares the same state, registration, and random seed as our full model. An additional control grid contains eight copies of the unedited source photograph. Both participants and the VLM judge choose A or B for each of the following questions.
\begin{itemize}
  \item \emph{Rotation:} Which group turns the object to the direction marked under each image more accurately?
  \item \emph{Consistency:} Which group looks more like photos of one and the same real object from different directions?
  (Consider shape, proportions, colours, patterns and details, including the sides hidden in the input.
  A group that barely turns the object, with near-identical images, counts as worse.)
\end{itemize}

\noindent\textbf{User study.}
We recruited 21 volunteers to complete this assessment. Each participant completed 24 comparisons, three against each alternative, and two attention checks using the unedited control. For each participant, source images were randomly selected without repetition, and the 26 comparisons were presented in random order. Our results appeared equally often as A and B, and method names were hidden. All participants passed both attention checks, so no responses were excluded. Finally, each baseline received 63 judgments per question. We report the percentage of judgments preferring our method.

\noindent\textbf{VLM judge.}
The VLM metrics in the main paper use Qwen3.8-27B to evaluate individual editing results, following the evaluation settings of the public benchmarks. Our cross-view evaluation instead requires comparing two groups of eight images. We initially tested Qwen3.8-27B on some source images, but found that it also assigned high consistency scores to groups with little viewpoint change, including repeated copies of the unedited source photograph. Such scores do not necessarily reflect consistency across different viewpoints, as nearly identical images can appear highly consistent without performing the requested transformations. We therefore used \texttt{gpt-6.1-sol} with high reasoning effort for the cross-view comparisons.

For each source, the judge evaluated all eight alternatives and the unedited control in both A/B orders, yielding $61 \times 9 \times 2 = 1{,}098$ evaluations. For each question, agreement across orders counts as a win for the preferred method, and disagreement counts as a tie. We report our preference rate against each alternative as $(\mathrm{wins}+\tfrac{1}{2}\,\mathrm{ties})/61$.

\begin{table}[t]
  \centering
  \small
  \setlength{\tabcolsep}{3pt}
  \resizebox{\columnwidth}{!}{%
  \begin{tabular}{@{}lcccc@{}}
    \toprule
    \multirow{2}{*}[-0.9ex]{\textbf{Ours} vs.}
      & \multicolumn{2}{c}{\textbf{User study (21)}}
      & \multicolumn{2}{c}{\textbf{VLM judge}} \\
    \cmidrule(lr){2-3}\cmidrule(lr){4-5}
      & Rotation & Consistency & Rotation & Consistency \\
    \midrule
    FLUX.1 Kontext          & 96.8 & 93.7 & 100  & 97.5 \\
    Qwen-Image-Edit-2511    & 77.8 & 68.3 & 90.2 & 69.7 \\
    SpatialEdit             & 61.9 & 68.3 & 54.9 & 57.4 \\
    TiB                     & 95.2 & 92.1 & 98.4 & 93.4 \\
    GeoDiffuser             & 98.4 & 98.4 & 100  & 100  \\
    Diffusion Handles       & 100  & 100  & 100  & 100  \\
    FreeFine                & 98.4 & 98.4 & 99.2 & 100  \\
    \midrule
    Paired training       & 58.7 & 61.9 & 55.7 & 53.3 \\
    Unedited control      & 100  & 100  & 100  & 100  \\
    \bottomrule
  \end{tabular}%
  }
  \caption{\textbf{Cross-view consistency evaluation.} We report the preference rates (\%) for our method in pairwise comparisons on 61 rotation sources from SpatialEdit-Bench.}
  \label{tab:supp_xview}
\end{table}

\noindent\textbf{Results.}
\Cref{tab:supp_xview} shows similar preference patterns between the user study and the VLM judge. Against FLUX.1 Kontext, TiB, GeoDiffuser, Diffusion Handles, and FreeFine, our method is preferred in over 92\% of comparisons on rotation and consistency by both raters. Our method is also preferred over Qwen-Image-Edit-2511 on both criteria. Among the external baselines, SpatialEdit is the strongest competitor in this evaluation, with lower preference rates for ours: 61.9\% and 68.3\% from the user study, and 54.9\% and 57.4\% from the VLM judge, for rotation and consistency, respectively. Both raters prefer our results over the unedited source in all control comparisons.

Compared with paired training, our method shows relatively modest gains, with participant preference rates of 58.7\% for rotation and 61.9\% for consistency, and VLM preference rates of 55.7\% and 53.3\%, respectively. This is expected because both methods use the same reconstructed 3D state to generate all eight views, which already provides the shape and appearance of the object from different viewpoints.
Moreover, the state is reconstructed from the source photograph, so the two sources of appearance largely agree on these benchmark images. The modest preference for our method may therefore be mainly attributed to its better image fidelity, consistent with the higher PSNR and lower LPIPS reported in Table~3 of the main paper.


\section{State Adherence on Real Photographs}
\label{sec:supp_state_adherence}

On visible surfaces, the reconstructed 3D state generally agrees with the source photograph. On hidden surfaces, a plausible edit alone does not reveal whether the editor followed the state. We therefore perturb the state colours separately on source-hidden and source-visible surfaces and measure how the output responds in each region.

Both tests use the same 61 source photographs from the rotation subset of SpatialEdit-Bench. For each edit, we invert the state colours on the selected surfaces ($S^{\prime c}_p = 1 - S^c_p$), leaving all other colours and state normals unchanged. We generate outputs $y$ and $y'$ using the original state $S$ and the modified state $S'$, respectively, with the same source photograph, reconstructed geometry, registration, and sampling noise.

Let $M$ denote the target-view pixels on the perturbed surfaces. We measure the component of the output colour change along the imposed state colour change as
\begin{equation}
  A(M) =
  \frac{\sum_{p \in M}
    \langle y'_p - y_p,\, S^{\prime c}_p - S^c_p \rangle}
       {\sum_{p \in M}
    \lVert S^{\prime c}_p - S^c_p \rVert^2}.
\end{equation}
We express this coefficient as a percentage. A score of 100\% indicates that the output matches the state change in the measured direction over $M$, while 0\% indicates no response in that direction.

We evaluate four cumulative variants. Starting from \emph{Paired training}, we successively add counterfactual supervision (\emph{CF supervision}), state-side correction (\emph{State correction}), and photo-side correction (\emph{Ours}).

\noindent\textbf{Hidden-side adherence (HSA).}
For each source, we select the two rotations with the largest absolute yaw among those with $|\mathrm{yaw}| \geq 120^\circ$. These 122 edits expose surfaces hidden in the source photograph. We compute HSA as $A(M)$ over the newly visible pixels on those surfaces and report the mean across the 61 sources.

\noindent\textbf{Visible-side adherence (VSA).}
For each source, we select the two rotations whose absolute yaw is closest to $40^\circ$ within the range $15^\circ \leq |\mathrm{yaw}| \leq 60^\circ$. These 122 edits have a mean absolute yaw of $41^\circ$ and retain much of the originally visible surface. We compute VSA as $A(M)$ over pixels on surfaces clearly visible in the source photograph, excluding regions near object boundaries and surfaces facing away from the source camera. A VSA near 0\% indicates that the imposed state colour change has little effect on visible-surface colours in the measured direction.

\begin{table}[!t]
  \centering
  \small
  \setlength{\tabcolsep}{3pt}
  \begin{tabular*}{\columnwidth}{@{\extracolsep{\fill}}lcccc@{}}
    \toprule
    \multirow{2}{*}[-0.9ex]{\textbf{Variant}}
    & \multicolumn{2}{c}{\textbf{HSA} (\%) $\uparrow$}
    & \multicolumn{2}{c}{\textbf{VSA} (\%) $\downarrow$} \\
    \cmidrule(lr){2-3}\cmidrule(lr){4-5}
    & Mean & Gain & Mean & Reduction\\
    \midrule
    Paired training  & 41.6 & --    & 11.0 & -- \\
    CF supervision   & 50.4 & +21.2 & 3.4 & 69.1 \\
    State correction & \textbf{58.4} & \textbf{+40.4} & 5.8 & 47.3 \\
    Ours             & 57.8 & +38.9 & \textbf{2.6} & \textbf{76.4} \\
    \bottomrule
  \end{tabular*}
  \caption{\textbf{Hidden- and visible-side adherence on real photographs.}
  HSA measures the response to state changes on hidden surfaces
  (higher is better), while VSA measures the influence of state
  changes on visible surfaces (lower is better).
  }
  \label{tab:supp_hsa}
  \par\vspace{7pt}
  \begin{minipage}{\columnwidth}
  \captionsetup{type=figure}
  \centering
  \includegraphics[width=\columnwidth]{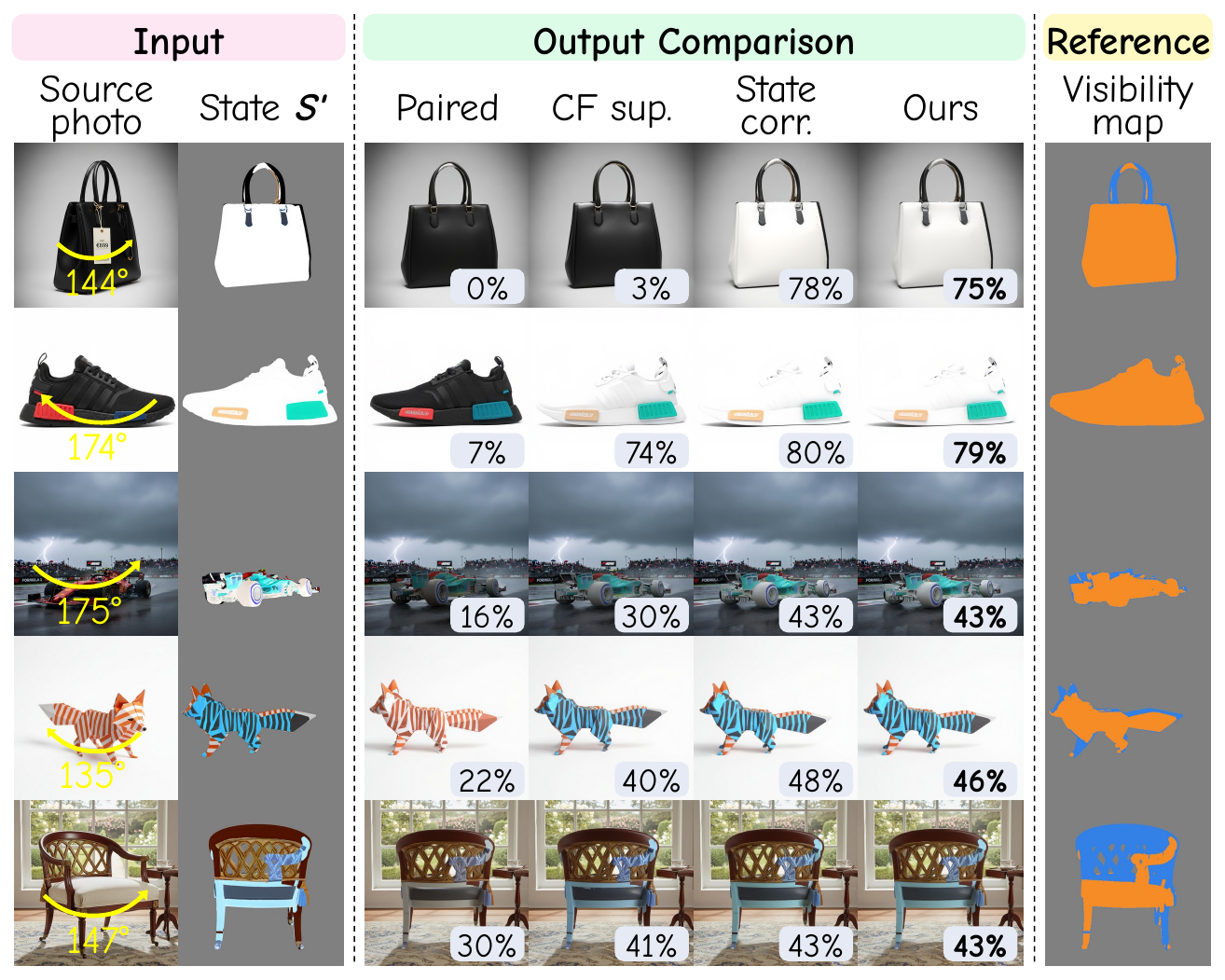}
  \caption{\textbf{Response to hidden-surface colour edits on real photographs.}
  The visibility map marks hidden surfaces in orange and visible surfaces in blue, both relative to the source view. Each output is labelled with its hidden-side adherence (HSA) score.}
  \label{fig:supp_hsa}
  \end{minipage}
\end{table}

\noindent\textbf{Results.}
As shown in \cref{tab:supp_hsa}, counterfactual supervision improves HSA from 41.6\% to 50.4\%, and state-side correction further increases it to 58.4\%. The full model achieves 57.8\%, remaining close to the state-side correction variant and outperforming paired training by 16.2 percentage points. Adding photo-side correction brings no further improvement in HSA, which is consistent with its role in preserving photograph-visible appearance rather than modifying hidden surfaces.

For VSA, counterfactual supervision reduces the score from 11.0\% to 3.4\%, indicating better preservation of visible appearance despite conflicting state colours. Adding state-side correction increases VSA to 5.8\%, suggesting that some state appearance leaks onto visible surfaces. Photo-side correction reduces VSA to 2.6\%, while largely preserving the improvement in HSA. These results highlight the complementary roles of the two correction strategies: state-side correction strengthens adherence to the state on hidden surfaces, while photo-side correction helps preserve the photographed appearance on visible surfaces.

The qualitative results in \cref{fig:supp_hsa} further illustrate these differences. With paired training, the model often retains the original colours despite changes to the hidden-side state, as seen in the black handbag and sneaker and the orange-striped paper fox. Counterfactual supervision improves the response in some cases, while state-side correction further strengthens it, particularly for the handbag. The full model largely preserves these improvements.
Together with the cross-view evaluation in \cref{sec:supp_xview}, these results show the importance of learning which conditioning source to trust, particularly when the 3D state provides information unavailable in the source photograph.


\section{Limitations}
\label{sec:supp_limitations}

Our method relies on a 3D asset reconstructed from a single photograph to guide spatial editing. Its performance can therefore be affected by reconstruction quality, particularly for large rotations that reveal surfaces not visible in the source photograph. Inaccurate geometry or textures on these surfaces may lead to errors in the edited results, even when the editor correctly follows the provided state. While our counterfactual training helps the editor preserve visible appearance and follow the state for unseen content, it cannot fully compensate for inaccuracies in the reconstructed 3D asset. Future work could incorporate reconstruction uncertainty to adapt how strongly the editor relies on the 3D state in newly revealed regions.

\makeatletter
\setlength{\@fptop}{0pt}
\makeatother
\end{document}